\documentclass[11pt]{article}
\usepackage{coling2020}
\usepackage{times}
\usepackage{url}
\usepackage{latexsym}

\colingfinalcopy 

\usepackage{color}
\usepackage{xcolor}
\definecolor{ugreen}{rgb}{0,0.5,0}
\definecolor{lgreen}{rgb}{0.9,1,0.8}
\definecolor{lightgray}{gray}{0.85}
\definecolor{myblack}{rgb}{0.15,0.15,0.15}
\definecolor{lyellow}{rgb}{0.54, 0.25, 0.27}
\usepackage{caption}
\usepackage{enumitem}
\usepackage{amsmath}
\usepackage{amsfonts} 
\usepackage{pgfplots}
\usepackage{tikz}
\usepackage{subfig}
\usetikzlibrary{backgrounds,fit} 
\usetikzlibrary{shapes,arrows,shadows}
\usetikzlibrary{patterns}
\usetikzlibrary{shapes.geometric} 
\usetikzlibrary{decorations.pathreplacing} 
\usetikzlibrary{calc}
\usepackage{array,multirow}
\usepackage{array} 
\usepackage{booktabs} 
\usepackage{arydshln} 
\newcommand{\PreserveBackslash}[1]{\let\temp=\\#1\let\\=\temp} 
\newcolumntype{C}[1]{>{\PreserveBackslash\centering}p{#1}}
\newcolumntype{R}[1]{>{\PreserveBackslash\raggedleft}p{#1}}
\newcolumntype{L}[1]{>{\PreserveBackslash\raggedright}p{#1}}
\newcolumntype{M}[1]{ >{\centering\arraybackslash}m{#1}}
\usepackage{cleveref}
\crefname{section}{§}{§§}
\Crefname{section}{§}{§§}

\pgfplotsset{compat=1.16}

\usepackage{verbatim}
\usepackage{xspace}

\newlength{\vseg}
\newlength{\hseg}
\newlength{\wnode}
\newlength{\hnode}
\newcommand{\encsa}[1]{H^{(#1)}}
\newcommand{\encsb}[1]{\dot{H}^{(#1)}}
\newcommand{\encfinal}{H^*}
\newcommand{\auxencfinal}{H_a^*}

\newcommand{\decsa}[1]{Z^{(#1)}}
\newcommand{\decsb}[1]{\dot{Z}^{(#1)}}
\newcommand{\decsc}[1]{\ddot{Z}^{(#1)}}
\newcommand{\decfinal}{Z^*}

\newcommand{\auxdecsb}[1]{\dot{Z}_a^{(#1)}}
\newcommand{\auxdecsc}[1]{\ddot{Z}_a^{(#1)}}

\newcommand{\citet}{\newcite}

\title{Layer-Wise Multi-View Learning for Neural Machine Translation}

\author{
	Qiang Wang$^1$\thanks{\xspace\xspace Work done during Ph.D. study at Northeastern University.},
	Changliang Li$^{2}$, 
	Yue Zhang$^{3}$\thanks{\xspace\xspace Corresponding author.}, 
	Tong Xiao$^{4,5}$,
	Jingbo Zhu$^{4,5}$  \\
	$^{1}$Machine Intelligence Technology Lab, Alibaba DAMO Academy, Hangzhou, China \\ 
	$^{2}$Kingsoft AI Lab, Beijing, China \\
	$^{3}$Institute of Advanced Technology, Westlake Institute for Advanced Study, Hangzhou, China\\
	$^{4}$ School of Computer Science and Engineering, Northeastern University, Shenyang, China \\
	$^{5}$NiuTrans Research, Shenyang, China \\ 
	{\tt
		zhiniao.wq@alibaba-inc.com,  lichangliang@kingsoft.com
	}\\
	{\tt
	
	zhangyue@westlake.edu.cn, 	\{xiaotong,zhujingbo\}@mail.neu.edu.cn
	} \\
	
}

\date{}

\begin{document}
\maketitle
\begin{abstract}
  Traditional neural machine translation is limited to the topmost encoder layer's context representation and cannot directly perceive the lower encoder layers. Existing solutions usually rely on the adjustment of network architecture, making the calculation more complicated or introducing additional structural restrictions. In this work, we propose layer-wise multi-view learning to solve this problem, circumventing the necessity to change the model structure. 
  We regard each encoder layer's off-the-shelf output, a by-product in layer-by-layer encoding, as the redundant view for the input sentence.
  In this way, in addition to the topmost encoder layer (referred to as the primary view), we also incorporate an intermediate encoder layer as the auxiliary view. 
  We feed the two views to a partially shared decoder to maintain independent prediction.  
  Consistency regularization based on KL divergence is used to encourage the two views to learn from each other.
  Extensive experimental results on five translation tasks show that our approach yields stable improvements over multiple strong baselines. As another bonus, our method is agnostic to network architectures and can maintain the same inference speed as the original model.  
\end{abstract}

\section{Introduction}

	\blfootnote{
	%
	%
	%
	%
	%
	%
	\hspace{-0.65cm}  
	This work is licensed under a Creative Commons 
	Attribution 4.0 International License.
	License details:
	\url{http://creativecommons.org/licenses/by/4.0/}
}

\noindent Neural Machine Translation (NMT) adopts the encoder-decoder paradigm to model the entire translation process \cite{bahdanau2015neural}. Specifically, the encoder finds a multi-layer representation of the source sentence, and the decoder queries the topmost encoding representation to produce the target sentence through a cross-attention mechanism \cite{wu2016google,vaswani2017attention}. However, such over-reliance on the topmost encoding layer is problematic in two aspects: (1) Prone to over-fitting, especially when the encoder is under-trained, such as in low-resource tasks \cite{wang2018multi}; (2) It cannot make full use of representations extracted from lower encoder layers, which are syntactically and semantically complementary to higher layers  \cite{peters2018deep,raganato-tiedemann-2018-analysis}. 

Researchers have proposed many methods to make the model aware of various encoder layers besides the topmost to mitigate this issue. Almost all of them resort to the adjustment of network structure, which can be further divided into two categories. The first is to merge the feature representations extracted by distinct encoder layers before being fed to the decoder \cite{wang2018multi,dou2018exploiting,wang-etal-2019-exploiting}. The differences between them lie in the design of the merge function: through self-attention \cite{wang2018multi}, recurrent neural network \cite{wang-etal-2019-exploiting}, or tree-like hierarchical merge \cite{dou2018exploiting}. Moreover, the second makes each decoder layer explicitly align to a parallel encoder layer \cite{he2018layer} or all encoder layers \cite{bapna2018training}.  However, the above methods either complicate the original model \cite{wang2018multi,dou2018exploiting,wang-etal-2019-exploiting,bapna2018training} or limit the model's flexibility, such as requiring the number of the encoder layers to be the same as the decoder layers \cite{he2018layer}.

Instead, in this work, we propose layer-wise multi-view learning to address this problem from the perspective of model training, without changing the model structure. Our method's highlight is that only the training process is concerned, while the inference speed is guaranteed to be the same as that of the standard model. The core idea is that we regard the off-the-shelf output of each encoding layer as a view for the input sentence. Therefore, it is straightforward and cheap to construct multiple views during a standard layer-by-layer encoding process. 
Further, in addition to the output of the topmost encoder layer used in standard models (refer to the \textit{primary view}), we also incorporate an intermediate encoder layer as the \textit{auxiliary view}. We feed the two views to a partially shared decoder for independent predictions. An additional regularization loss based on prediction consistency between views is used to encourage the auxiliary view to mimic the primary view. Thanks to the co-training on the two views, the gradients during back-propagation can simultaneously flow into the two views, which implicitly realizes the knowledge transfer.

Extensive experimental results on five translation tasks (Ko$\rightarrow$En, IWSLT'14 De$\rightarrow$En, WMT'17 Tr$\rightarrow$En, WMT'16 Ro$\rightarrow$En, and WMT'16 En$\rightarrow$De) show that our method can stably outperform multiple baseline models \cite{vaswani2017attention,wang2018multi,dou2018exploiting,bapna2018training}. In particular, we have achieved new state-of-the-art results of 10.8 BLEU on Ko$\rightarrow$En and 36.23 BLEU on IWSLT'14 De$\rightarrow$En. Further analysis shows that our method's success lies in the robustness to encoding representations and dark knowledge \cite{hinton2015distilling} provided by consistency regularization.

\section{Approach}

\noindent In this section, we will take the Transformer model \cite{vaswani2017attention} as an example to show how to train a model by our multi-view learning. We first briefly introduce Transformer in \cref{sec:transformer}, then describe the proposed multi-view Transformer model (called \texttt{MV-Transformer}) and its training and inference in detail in \cref{sec:mv_transformer}. Finally, we discuss why our method works in \cref{sec:mv_discuss}. See Figure \ref{fig:model} for an overview of the proposed approach. 

\subsection{Transformer}
\label{sec:transformer}

\noindent 
The Transformer systems follow the encoder-decoder paradigm. On the encoder side, there are $M$ identical stacked layers. Each of them comprises a self-attention-network (SAN) sub-layer and a feed-forward-network (FFN) sub-layer. To easy optimization, layer normalization (LN) \cite{ba16layer} and residual connections \cite{he2016deep} are used between these sub-layers. There are two ways to incorporate them, namely \texttt{PreNorm} Transformer and \texttt{PostNorm} Transformer \cite{wang2019learning}. 
Without loss generalization, here we only describe the implementation of PreNorm Transformer, but we test our method in both cases\footnote{The basic form of PostNorm Transformer is: $y=\textrm{LN}(x + \textrm{SubLayer}(x))$, and there is no additional layer normalization on the top of encoder/decoder.}. The $l$-th encoder layer of PreNorm Transformer is:
\begin{eqnarray}
\label{eq:post_enc}
\begin{split}
\encsb{l} &= \encsa{l-1} + \textrm{SAN}\big(\textrm{LN}(\encsa{l-1})\big) \\
\encsa{l} &= \encsb{l} + \textrm{FFN}\big(\textrm{LN}(\encsb{l})\big) 
\end{split}
\end{eqnarray} 

\noindent 
where $\dot{H}$ denotes the intermediate encoding state after the first sublayer. Besides, there is an extra layer normalization behind the topmost layer to prevent the excessive accumulation of the unnormalized output in each layer, i.e.
$\encfinal=\textrm{LN}(\encsa{M})$, where $\encfinal$ denotes the final encoding result. 
\noindent Likewise, the decoder has another stack of $N$ identical layers, but an additional cross-attention-network (CAN) sub-layer is inserted between SAN and FFN compared to the encoder layer:
\begin{eqnarray}
\label{eq:post_dec}
\begin{split}
\decsc{l} &= \decsb{l} + \textrm{CAN}\big(\textrm{LN}(\decsb{l}), \encfinal\big)
\end{split}
\end{eqnarray} 

\noindent $\textrm{CAN}(\cdot)$ is similar to $\textrm{SAN}(\cdot)$ except that its \textit{key} and \textit{value} are composed of the encoding output $\encfinal$ instead of \textit{query} itself. Resemble $\encfinal$, the last extracted feature vector by decoder is $\decfinal=\textrm{LN}(\decsa{N})$. Thus, given a sentence pair of $\langle \textbf{x}, \textbf{y} \rangle$, where $\textbf{x}=(x_1,\ldots,x_{m})$ and $\textbf{y}=(y_1,\ldots,y_{n})$, we can train the model parameters $\theta$ by minimizing the negative log-likelihood:
\begin{equation}
\label{eq:mle}
\mathcal{L}_{nll}(\theta) = -\sum_{j=1}^{n}{\textrm{log}p_{\theta}(y_j|\textbf{x}, y_{<j})}
\end{equation} 
\noindent
where $p_{\theta}(y_j|\textbf{x}, y_{<j})=\textrm{Softmax}(W_o \decfinal_j + b_o)$, $W_o$ and $b_o$ are the parameters in the output layer.

\subsection{Multi-View Transformer}
\label{sec:mv_transformer}

\begin{figure}[t]
	\begin{center}
		\begin{tikzpicture}[decoration=brace, scale=0.9,
		triangle/.style = {regular polygon, regular polygon sides=3, inner sep=0.5pt},]
		\begin{scope}
		
		\setlength{\vseg}{3em}
		\setlength{\wnode}{3em}
		\setlength{\hnode}{2em}
		\newlength{\vpoint}
		\setlength{\vpoint}{7pt}
		\newlength{\hpoint}
		\setlength{\hpoint}{2pt}
		\setlength{\hseg}{0.7\vseg}
		
		\tikzstyle{newynode} = [draw opacity=0, fill=green!20, draw, circle, thin, inner sep=1pt, minimum size=0.3\hnode]
		\tikzstyle{oldynode} = [draw opacity=0,fill=green!20, draw, circle, thin, inner sep=1pt, minimum size=0.3\hnode]
		\tikzstyle{weightnode} = [draw opacity=0, fill=blue!20, draw, circle, thin, inner sep=1pt, minimum size=0.3\hnode, anchor=west]
		
		\tikzstyle{layernode} = [draw, thin, rounded corners=1pt, inner sep=1pt, fill=yellow!20, minimum height=0.6\hnode, minimum width=1.3\wnode]
		\tikzstyle{lnnode} = [draw, thin, rounded corners=1pt, inner sep=1pt, fill=yellow!20, minimum height=.5\hnode, minimum width=.8\wnode]
		\tikzstyle{declayernode} = [draw, thin, dashed, rounded corners=3pt, inner sep=1pt, minimum height=3\hnode, minimum width=\wnode]
		\tikzstyle{sublayernode} = [draw, thin, rounded corners=1pt, inner sep=1pt, fill=yellow!20, minimum height=0.5\hnode, minimum width=0.8\wnode]
		\tikzstyle{legendnode} = [thin, rounded corners=1pt, inner sep=1pt, minimum height=0.3\hnode, minimum width=0.5\wnode]

		\node [] (input) at (0,0) {\scriptsize{Input}};
		
		\node[layernode, anchor=north, fill=yellow!15] (enc-l1) at ([yshift=.8\vseg]input.north) {\scriptsize{Layer 1}};
		\draw [->, ] (input) to node [auto] {} (enc-l1)  ;
		\node [inner sep=0pt] (enc-l1-mid) at ([yshift=0.07\vseg]enc-l1.north) {\scriptsize{$\cdots$}};
		
		\node[layernode, anchor=north, fill=yellow!15] (enc-la) at ([yshift=.9\vseg]enc-l1.north) {\scriptsize{{Layer $M_a$}}};
		\node[lnnode, anchor=west, fill=blue!20, label=\scriptsize{$\auxencfinal$}] (enc-la-ln) at ([xshift=-2.7\hnode,yshift=0.3\vseg]enc-la.north) {\scriptsize{{LN}}};
		\draw [->, very thin ] ([yshift=.05\vseg]enc-la.north) .. controls +(-0.5,+.5) and +(.5,-.3) .. (enc-la-ln.east);
		
		\node[layernode, anchor=north, fill=red!20] (enc-la-next) at ([yshift=.8\vseg]enc-la.north) {\scriptsize{Layer $M_a$+1}};
		\node [inner sep=0pt] (enc-la-mid) at ([yshift=0.07\vseg]enc-la-next.north) {\scriptsize{$\cdots$}};
		\draw [->, ] (enc-la) to node [auto] {
		} (enc-la-next)  ;

		\node[layernode, anchor=north, fill=red!20] (enc-le) at ([yshift=.7\vseg]enc-la-next.north) {\scriptsize{Layer $M$}};
		\node [] (enc-top) at ([yshift=0.6\vseg]enc-le.north) {};
		\draw [->, ] (enc-le) to node [auto] {} (enc-top)  ;
		\node [inner sep=0pt] (enc-le-mid) at ([yshift=0.25\vseg]enc-le.north) {};
		\node[lnnode, anchor=east, fill=red!20, label=below:\scriptsize{$\encfinal$}] (enc-le-ln) at ([xshift=2.7\hnode]enc-le-mid.east) {\scriptsize{{LN}}};
		\draw [->, very thin] ([yshift=.1\vseg]enc-le.north) .. controls +(.5,.5) and +(-.5,-.5) ..  (enc-le-ln.west)  ;
		\node[fit=(input)(enc-le-ln)(enc-le)(enc-la-ln), draw, very thin, densely dashed,rounded corners=5pt, black!40, label=\scriptsize{\textbf{Encoder}}] {};
		
		\node [] (pri-input) at ([xshift=5\hnode]input.east) {\scriptsize{Primary}};
		\node[sublayernode, anchor=north, fill=yellow!15] (pri-sub1) at ([yshift=0.5\vseg]pri-input.north) {\scriptsize{{SA}}};
		\node[sublayernode, anchor=north, fill=red!20] (pri-sub2) at ([yshift=0.6\vseg]pri-sub1.north) {\scriptsize{{CA}}};
		\node[sublayernode, anchor=north, fill=yellow!15] (pri-sub3) at ([yshift=0.5\vseg]pri-sub2.north) {\scriptsize{{FFN}}};
		\draw [->, ] (pri-sub1) to node [auto] {} (pri-sub2)  ;
		\draw [->, ] (pri-sub2) to node [auto] {} (pri-sub3)  ;
		
		\node[sublayernode, anchor=north, fill=yellow!15] (pri-ln) at ([yshift=0.73\vseg]pri-sub3.north) {\scriptsize{{LN}}};
		\node[sublayernode, anchor=north, fill=yellow!15] (pri-wo) at ([yshift=0.5\vseg]pri-ln.north) {\scriptsize{{$W_o$}}};
		\draw [->, ] (pri-sub3) to node [auto] {} (pri-ln)  ;
		\draw [->, ] (pri-ln) to node [auto] {} (pri-wo)  ;
		\node[anchor=north] (pri-nll) at ([yshift=0.8\vseg]pri-wo.north) {\scriptsize{{$\mathcal{L}_{\textrm{nll}}^{\textrm{pri}}$}}};
		\draw [->, ] (pri-wo) to node [auto] {} (pri-nll)  ;
		\node[inner sep=0pt] (pri-context) at ([xshift=-0.9\vseg, yshift=-0.1\vseg]pri-sub1.north) {\scriptsize{$\encfinal$}};
		\draw [->, semithick, densely dotted, red!40] (pri-context.east) .. controls +(north:0.4cm) and +(0.1cm, -0.4cm) .. ([xshift=-4pt]pri-sub2.south)  ;
		\node[fit=(pri-sub1)(pri-sub2)(pri-sub3), draw, densely dashed, thin, black!70, rounded corners=3pt, inner sep=5pt, label={[xshift=-0.9cm,yshift=-0.4cm]\tiny{$ \textrm{N} \times$}}] {};
		
		\node [] (aux-input) at ([xshift=1.5\hnode]pri-input.east) {\scriptsize{Auxiliary}};
		\node[sublayernode, anchor=north, fill=yellow!15] (aux-sub1) at ([yshift=0.5\vseg]aux-input.north) {\scriptsize{{SA}}};
		\node[sublayernode, anchor=north, fill=blue!20] (aux-sub2) at ([yshift=0.6\vseg]aux-sub1.north) {\scriptsize{{CA}}};
		\node[sublayernode, anchor=north, fill=yellow!15] (aux-sub3) at ([yshift=0.5\vseg]aux-sub2.north) {\scriptsize{{FFN}}};
		\draw [->, ] (aux-sub1) to node [auto] {} (aux-sub2)  ;
		\draw [->, ] (aux-sub2) to node [auto] {} (aux-sub3)  ;
		\node[fit=(aux-sub1)(aux-sub2)(aux-sub3), draw, densely dashed,rounded corners=3pt, inner sep=5pt, thin, black!70, label={[xshift=0.9cm,yshift=-0.4cm]\tiny{$\times \textrm{N}$}}] {};
		\node[sublayernode, anchor=north, fill=yellow!15] (aux-ln) at ([yshift=0.73\vseg]aux-sub3.north) {\scriptsize{{LN}}};
		\node[sublayernode, anchor=north, fill=yellow!15] (aux-wo) at ([yshift=0.5\vseg]aux-ln.north) {\scriptsize{{$W_o$}}};
		\draw [->, ] (aux-sub3) to node [auto] {} (aux-ln)  ;
		\draw [->, ] (aux-ln) to node [auto] {} (aux-wo)  ;
		\node[anchor=north] (aux-nll) at ([yshift=0.73\vseg]aux-wo.north) {\scriptsize{{$\mathcal{L}_{\textrm{nll}}^{\textrm{aux}}$}}};
		\draw [->, ] (aux-wo) to node [auto] {} (aux-nll)  ;
		\node[inner sep=0pt] (aux-context) at ([xshift=0.9\vseg, yshift=-0.1\vseg]aux-sub1.north) {\scriptsize{$\auxencfinal$}};
		\draw [->, semithick, densely dotted, blue!40] (aux-context.west) .. controls +(north:0.4cm) and +(0.1cm, -0.4cm) .. ([xshift=4pt]aux-sub2.south)  ;
		
		\node[anchor=east] (kl) at ([xshift=0.73\vseg]pri-nll.east) {\scriptsize{{$\hat{\mathcal{L}}_{\textrm{kl}}$}}};
		\node [inner sep=0pt] (pri-top) at ([yshift=0.1\vseg]pri-wo.north) {};
		\node [inner sep=0pt] (aux-top) at ([yshift=0.1\vseg]aux-wo.north) {};
		\draw[->,bend right=90, looseness=3] (pri-top.east) .. controls +(0.7,0) .. ([xshift=-2pt]kl.south);
		\draw[->,bend right=90, looseness=3] (aux-top.west) .. controls +(-0.7,0) .. ([xshift=2pt]kl.south);
		\node[fit=(pri-input)(pri-context)(kl)(aux-input)(aux-context), draw, very thin, densely dashed,rounded corners=5pt,  black!40, label=\scriptsize{\textbf{Decoder}}] {};
		
		
		\node[legendnode, anchor=east, fill=yellow!20, label=right:\scriptsize{Shared}] (legend-share) at ([xshift=3\hnode,yshift=1\vseg]aux-input.east) {};

		\node[legendnode, anchor=east, fill=red!20, label=right:\scriptsize{Primary Only}] (legend-pri) at ([yshift=-.3\vseg,xshift=.9em]legend-share.south) {};
		
		\node[legendnode, anchor=south, fill=blue!20, label=right:\scriptsize{Auxialiary Only}] (legend-aux) at ([yshift=-.4\vseg]legend-pri.south) {};
		
		\begin{pgfonlayer}{background}
		\path[fill=gray!10,rounded corners] ([xshift=-.2em,yshift=.2em]legend-share.north west) rectangle ([xshift=5.2em, yshift=-.2em]legend-aux.south east);
		\end{pgfonlayer}
		\node[fit=(legend-share)(legend-pri)(legend-aux), very thin, densely dashed,rounded corners=5pt, black!40, xshift=.8cm, label=\scriptsize{\textbf{Legend}}] {};

		\end{scope}
		\end{tikzpicture}\end{center}
	
	\begin{center}
		\vspace{-0.3em}
		\caption{An overview of proposed layer-wise multi-view learning for PreNorm Transformer. We maintain a two-stream decoder during training by partially sharing.}
		\label{fig:model}
		\vspace{-0.0em}
	\end{center}
\end{figure}
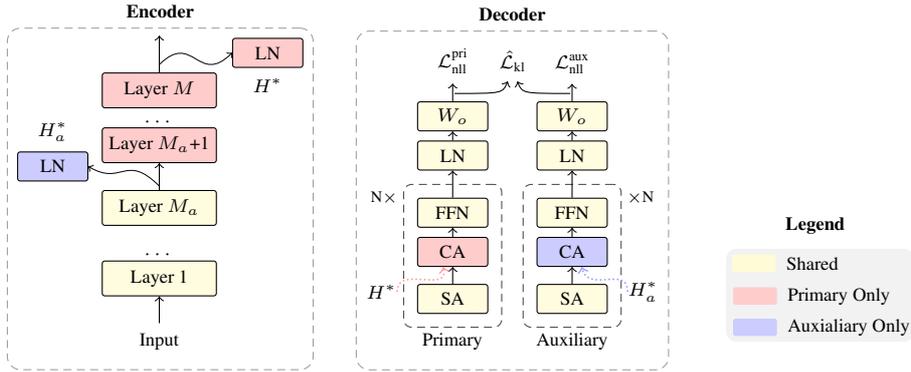

\begin{enumerate}[leftmargin=0pt]
	
	\item[] \textbf{Multi-view. }
	\noindent  Multi-view learning has achieved great success in conventional machine learning by exploiting the redundant views of the same input data \cite{xu2013survey}, where one of the keys is view construction. In our scenario, a view is the hidden representation of the input sentence (an array of hidden vectors for each token, e.g., $\encfinal$). In this work, we further propose to take the off-the-shelf output of each encoder layer (i.e., $\encsa{l}$ \footnote{To be precise, there is a newly added layer normalization following $\encsa{l}$ to keep consistent with $\encfinal$.}) to construct the redundant views.
	In NLP, previous implementations of view construction generally require the model to recalculate on the reconstructed input, such as using different orders of n-grams in the bag-of-word model \cite{Matsubara:2005:MSL:1565899.1565913}, randomly masking the input tokens \cite{clark-etal-2018-semi}. As opposed to them, our method is very cheap as the by-product of the standard layer-by-layer encoding process.
	According to the definition of a view, we can regard the vanilla Transformer as a single-view model since only the topmost encoder layer (also called \textit{primary view}) is fed to the decoder.
	In contrast, MV-Transformer additionally contains an intermediate layer $M_a$ ($1 \le M_a < M$) as the \textit{auxiliary view} \footnote{In this work, we only consider the case of one auxiliary view. More auxiliary views may be more helpful, while it requires more training costs. We leave this issue in future work.}. The choice of $M_a$ can be arbitrary, and we discuss its effect in \cref{sec:exp_hp}. Our goal is to learn a better single model with the help of the auxiliary view.
	
	\item[] \textbf{Partially shared parameters. } 
	\noindent In the encoder, except for the last layer normalization, all other parameters are shared in the two views to obtain the corresponding view representations by encoding only once. However, the situation is different for the decoder.
	We empirically find that a fully shared decoder has no sufficient capacity to be compatible with two different views simultaneously, especially in medium or large translation tasks (see \cref{sec:exp_ablation}). 
	On the other hand, it can be seen that the difference between the two views only directly affects the CANs in the decoder and has nothing to do with other sublayers (i.e., SANs, FFNs). Therefore, using a separate decoder for each view will cause an enormous waste of decoder model parameters.
	To trade-off, we extend the decoder network by using independent CANs for each view but share all SANs and FFNs (see Figure~\ref{fig:model}) \footnote{In earlier experiments, we tried to add \textit{view embedding} to make the decoder aware of identity of each view, i.e. $\encfinal + \mathcal{E}_{pri}$ for primary view, where $\mathcal{E}_{pri} \in \mathcal{R}^{d}$, but it did not work well.}.
	
	\item[] \textbf{Two-stream decoder. } 
	Given the two views, we use a two-stream decoder during training. Like Eq.~\ref{eq:post_dec}, the auxiliary view is queried as:
	\begin{equation}
	\label{eq:post_dec_aux}
	\auxdecsc{l} = \auxdecsb{l} + \textrm{CAN}_a\big(\textrm{LN}_a(\auxdecsb{l}), \auxencfinal\big)
	\end{equation} 
	\noindent where the subscript $a$ indicates used for auxiliary view. In each decoding step, one stream queries $\decsb{l}$ from the primary view $\encfinal$ like a standard Transformer, while the other stream queries $\auxdecsb{l}$ from the auxiliary view $\auxencfinal$ through separate CAN sublayers. 
	In this way, each stream yields distinct predictions based on different context semantics in the views. Here we use $p_{pri}(\cdot)$ and $p_{aux}(\cdot)$ to denote the prediction distribution by the primary view and the auxiliary view, respectively. 
	
	\item[] \textbf{Training. } 
	To jointly train the two views and transfer the knowledge between them, the training objective of MV-Transformer consists of two items. The first item $\hat{\mathcal{L}}_{nll}$ is similar to the negative log-likelihood in Eq.~\ref{eq:mle}, but additionally considers the log-likelihood of the auxiliary view prediction:
	\begin{equation}
		\label{eq:mv_mle}
		\hat{\mathcal{L}}_{nll} = \frac{1}{2} \times (\mathcal{L}_{nll}^{pri} + \mathcal{L}_{nll}^{aux})
	\end{equation} 
	\noindent where $\mathcal{L}_{nll}^{pri}$, $\mathcal{L}_{nll}^{aux}$ are based on the distribution of $p_{pri}(\cdot)$ and $p_{aux}(\cdot)$ respectively, and $1/2$ is used to numerically scale $\hat{\mathcal{L}}_{nll}$ to $\mathcal{L}_{nll}$.
	The second item $\hat{\mathcal{L}}_{cr}$ is the consistency regularization loss between views, where we use Kullback–Leibler (KL) divergence to let the student (played by the auxiliary view) imitate the prediction of the teacher (played by the primary view):
	\begin{eqnarray}
		\label{eq:mv_kl}
		\begin{split}
			\hat{\mathcal{L}}_{cr} &= -\sum_{j=1}^{n} \textrm{KL}(p_{aux}^{(j)}||p_{pri}^{(j)}) = -\sum_{j=1}^{n} \sum_{v \in \mathcal{V}} {p_{pri}^{(j)}(v) \textrm{log} \frac{p_{pri}^{(j)}(v)}{p_{aux}^{(j)}(v)}} 
		\end{split}
	\end{eqnarray} 
	\noindent where $p^{(j)}(v)$ is the probability of generating token $v$ at step $j$\footnote{We use temperature $\tau$=1 in all experiments, e.g. $p(i)=\textrm{exp}(z_i/\tau)/\sum_{j \in \mathcal{V}} \textrm{exp}(z_j/\tau)$, $z$ is the logit.}. 
	\noindent 
	We note that our consistency regularization is different from traditional knowledge distillation, where a typical implementation is to detach the teacher's prediction $p_{pri}^{(j)}$ as a constant \cite{hinton2015distilling}. On the contrary, our method takes $p_{pri}^{(j)}$ as a variable that requires gradients during back-propagation. To this end, the entire model parameters are optimized to give good predictions in two views instead of considering only one, which implicitly makes the model learn from different encoder layers.   
	Some people may say that it is enough for the student to learn from the teacher, but the reverse is unreasonable. However, we believe that the information in different views is complementary, so the potential for mutual learning of views may be greater than one-way learning. And our empirical comparison in \cref{sec:cmp_kd_ensemble} also confirms this assumption. 
	\noindent
	Finally, we can interpolate these two losses with the hyper-parameter $\alpha$ to obtain the overall loss function for multi-view learning:
	\begin{eqnarray}
		\label{eq:mv_loss}
		\hat{\mathcal{L}} = (1-\alpha) \times \hat{\mathcal{L}}_{nll} + \alpha \times \hat{\mathcal{L}}_{cr}
	\end{eqnarray} 
	\noindent Intuitively, when $\alpha$ is low, the loss degrades into Eq.~\ref{eq:mv_mle}, which only focuses on the ground-truth labels. On the contrary, a high $\alpha$ overemphasizes the consistency of the entire vocabulary between the two views, resulting in neglecting to learn from the provided ground-truth. 
	We discuss $\alpha$'s effect in ~\cref{sec:exp_hp}.
	
	\item[] \textbf{Inference. } 
	Instead of maintaining both views like training, we can shift to any single view at inference time. Considering the primary view as an example: We can straightforwardly discard all the modules attached to the auxiliary view, including $\textrm{CAN}_a$ and $\textrm{LN}_a$ in the decoder as well as the newly added layer normalization in the encoder. It makes the decoding speed to be precisely the same as that of the standard model. Likely, we can also switch to the auxiliary view composed of fewer encoder layers for slightly faster speed, but with the risk of performance degradation.  
	
\end{enumerate}

\subsection{Discussion}\label{sec:mv_discuss}

	\begin{figure}[t]
		\begin{center}
			\begin{tikzpicture}{baseline}
			\scriptsize{
				\begin{axis}[
				ylabel near ticks,
				width=.4\textwidth,
				height=.2\textwidth,
				legend style={at={(0.8, 0.5)}, anchor=north east},
				xlabel={$i$},
				ylabel={\scriptsize{Cosine Similarity [\%]}},
				ylabel style={yshift=-0em},xlabel style={yshift=0.0em},
				yticklabel style={/pgf/number format/precision=0,/pgf/number format/fixed zerofill},
				ymin=40,ymax=100, ytick={40, 50, 60, 70, 80, 90, 100},
				xmin=0,xmax=12, xtick={0, 1, 2, 3, 4, 5, 6, 7, 8, 9, 10, 11, 12},
				xmajorgrids,
				ymajorgrids,
				legend style={legend plot pos=left,cells={anchor=west}}
				]
				
				\addplot[blue,mark=otimes*,line width=0.5pt] coordinates {(0, 	45.26) (1, 64.67) (2, 73.64) (3, 78.46) (4, 83.09) (5, 87.24) (6, 89.89) (7, 92.41) (8, 94.66) (9, 96.29) (10, 97.97) (11, 99.24) (12, 100)};
				
				\end{axis}
			}
			\label{fig:enc_sim}
			\end{tikzpicture}
			
			\vspace{-0.5em}
			\caption{Cosine similarity between the $i$-th encoder layer and the topmost encoder layer in a pre-trained PreNorm Transformer with a 12-layer encoder. The results were measured on the IWSLT'14 De$\rightarrow$En validation set. $i$=0 indicates the embedding layer. }
			\label{fig:enc_sim}
			\vspace{-1.em}
			
		\end{center}

	\end{figure}
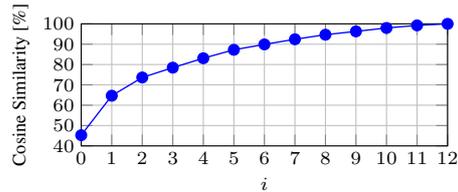
	
	In this section, we discuss why our method works from two aspects: robustness to encoding representation and dark knowledge. See ~\cref{sec:exp_why} for more experimental analysis. 

\begin{enumerate}[leftmargin=0pt]
	
	\item[] \textbf{Robustness to encoding representation. } Over-reliance on the top encoding layer (primary view) makes the model easier to over-fit \cite{wang2018multi}. Our method attempts to reduce the sensitivity to the primary view by feeding an auxiliary view. Figure~\ref{fig:enc_sim} shows that the vector similarity between the $i$-th encoder layer and the topmost layer grows as the increase of $i$. Therefore, we can regard the middle layer's auxiliary view as a noisy version of the primary view. Training with noises has been widely proven to effectively improve the model's generalization ability, such as dropout \cite{srivastava2014dropout}, adversarial training \cite{vat2017,cheng-etal-2019-robust} etc. We also experimentally confirm that our model is more robust than the single view model when injecting random noises into the encoding representation.
	
	\item[] \textbf{Dark knowledge. }  Typically, the prediction target in $\mathcal{L}_{nll}$ is a one-hot distribution: Only the gold label is 1, while the others are 0. A better alternative is label smoothing \cite{szegedy2016rethinking}, which reduces the probability of gold label by $\epsilon$ and redistributes $\epsilon$ to all non-gold labels on average. However, label smoothing ignores the relationship between non-gold labels. For example, if the current ground-truth is ``improve'', then ``promote'' should have high probability than ``eat''. In contrast, in our method, the target in the auxiliary view is the primary view's prediction, which contains more information about non-gold labels, also known as \texttt{dark knowledge} \cite{hinton2015distilling}. 
	
\end{enumerate}

\section{Experiments}

\subsection{Setup}
\begin{enumerate}[leftmargin=0pt]
	
	\item[] \textbf{Datasets. }  We conducted experiments on five translation tasks: Korean$\rightarrow$English (Ko$\rightarrow$En, 96k)\footnote{\url{https://sites.google.com/site/koreanparalleldata/}}, IWSLT'14 German$\rightarrow$English (De$\rightarrow$En, 160k), WMT'17 Turkish$\rightarrow$English (Tr$\rightarrow$En, 205k)
	, WMT'16 Romanian$\rightarrow$English (Ro$\rightarrow$En, 601k)\footnote{\url{https://github.com/nyu-dl/dl4mt-nonauto}} and WMT'16 English$\rightarrow$German (De$\rightarrow$En, 4.5M)\footnote{\url{https://drive.google.com/uc?export=download&id=0B_bZck-ksdkpM25jRUN2X2UxMm8}}. We use the officially provided development sets and test sets for all tasks. We pre-process and tokenize all the data sets using the Moses toolkit. 
	
	\item[] \textbf{Models and hyperparameters.}  We tested all models in shallow networks based on PostNorm and deep networks based on PreNorm,  respectively. Concretely, for shallow models, we use $M$=$N$=3 for the small-scale Ko-En task \footnote{We failed to train with $M$=$N$=6 in early experiments.}, while $M$=$N$=6 for other tasks. We use \textit{Small} configuration (embed=512, ffn=1024, head=4) for \{Ko, De, Tr\}-En, and \textit{Base} configuration (embed=512, ffn=2048, head=8) for Ro-En and En-De. As for deep models, we double the encoder depth as the corresponding PostNorm counterparts. E.g., suppose we use a 6-layer encoder in vanilla Transformer. In that case, we turn it to 12-layer in deep Transformer. For MV-Transformer, we use 1/3/6-th encoder layer as the auxiliary view when the encoder depth is 3/6/12, respectively.    
	Following \citet{vaswani2017attention}, we use the \textit{inverse\_sqrt} learning rate schedule with warm-up and label smoothing of 0.1. Some training hyperparameters are distinct across tasks due to the different data sizes.
	Detailed hyperparameters are listed in Appendix A.
	
	\item[] \textbf{Decoding and evaluation.}  To compare with previous works, we use the beam size of 4 and average last 5 checkpoints on De$\rightarrow$En, while for other tasks, we use the beam size of 5 and the best checkpoint according to the best BLEU score on the development set.
	For evaluation, except that Ko$\rightarrow$En uses \textit{sacrebleu} \footnote{BLEU+c.mixed+l.ko-en+\#.1+s.exp+tok.13a+v.1.2.21}, all other datasets are evaluated by \textit{multi-bleu.perl}. Only De$\rightarrow$En is reported by case insensitive BLEU. 
\end{enumerate}

\subsection{Main results}
\label{sec:exp_main}

\begin{table*}[t]
	\begin{center}
		\renewcommand\arraystretch{0.8}
		\begin{tabular}{l l| c  c  c  c  c}
			\toprule[1pt]
			
			\multicolumn{2}{c ||}{\textbf{Model}} &
			\multicolumn{1}{c }{\textbf{Ko$\rightarrow$En}} & 
			\multicolumn{1}{c }{\textbf{De$\rightarrow$En}} &
			\multicolumn{1}{c }{\textbf{Tr$\rightarrow$En}} &
			\multicolumn{1}{c }{\textbf{Ro$\rightarrow$En}} &
			\multicolumn{1}{c }{\textbf{En$\rightarrow$De}} \\
			
			\hline \hline
			\multicolumn{7}{c}{\textit{Shallow networks with PostNorm}} \\
			\hline
			\multicolumn{1}{l|}{\multirow{6}{*}{Aux.}} &
			\multicolumn{1}{c||}{Transformer$^\dagger$} 		
			& 8.7  & 34.10  	& 14.32  	& 33.07   & 32.80  \\ 
			\multicolumn{1}{l|}{} &
			\multicolumn{1}{c||}{MLRF$^\dagger$} 		
			& 9.1  & 34.73  	& \textbf{14.97}  	& 33.7  & 33.21 \\ 
			\multicolumn{1}{l|}{} & \multicolumn{1}{c||}{HieraAgg$^\dagger$} 		
			& N/A  & N/A  	& N/A  	& N/A  & N/A \\ 
			\multicolumn{1}{l|}{} & \multicolumn{1}{c||}{TA$^\dagger$} 		
			& 8.8  & 34.23  	& 14.51  	& 33.15  & 32.92 \\ 
			\multicolumn{1}{l|}{} & \multicolumn{1}{c||}{MV-Transformer} 		
			& \textbf{10.2}  & \textbf{35.25}  	& 14.74  	& \textbf{34.24}  & \textbf{33.38} \\ 
			\multicolumn{1}{l|}{} & \multicolumn{1}{c||}{$\Delta$} 		
			& +1.5 		& +1.15 	& +0.42 	& +1.17 & +0.58 \\ \hline 
			
			\multicolumn{1}{l|}{\multirow{6}{*}{Pri.}} &
			\multicolumn{1}{c||}{Transformer$^\dagger$} 		
			& 9.6  	& 34.77  	& 14.78  	& 33.20  & 33.06  \\ 
			\multicolumn{1}{l|}{} &
			\multicolumn{1}{c||}{MLRF$^\dagger$} 		
			& 10.0  & 33.53  	& 15.18  	& 33.79  & 33.17 \\ 
			\multicolumn{1}{l|}{} & \multicolumn{1}{c||}{HieraAgg$^\dagger$} 		
			& N/A  & 34.98  	& 14.75  	& 34.09  & 33.36 \\ 
			\multicolumn{1}{l|}{} & \multicolumn{1}{c||}{TA$^\dagger$} 		
			& 9.1  & 34.58  	& 15.06  	& 33.49  & 32.97 \\ 
			\multicolumn{1}{l|}{} & \multicolumn{1}{c||}{MV-Transformer} 		
			& \textbf{10.4}  & \textbf{35.49}  	& \textbf{15.25}  	& \textbf{34.45}$^*$  & \textbf{33.75}  \\ 
			\multicolumn{1}{l|}{} & \multicolumn{1}{c||}{$\Delta$} 		
			& +0.8 		& +0.72 		& +0.47 		& +1.25 & +0.69 \\ \hline \hline
			\multicolumn{7}{c}{\textit{Deep networks with PreNorm}} \\
			\hline
			
			\multicolumn{1}{l|}{\multirow{6}{*}{Aux.}} &
			\multicolumn{1}{c||}{deep Transformer$^\dagger$} 		
			& 9.1  	& 35.38  	& 14.27 		& 33.15 & 33.61 \\ 
			\multicolumn{1}{l|}{} &
			\multicolumn{1}{c||}{MLRF$^\dagger$} 		
			& 9.5  & 35.32 	& 14.71  	& 33.55  & 33.51 \\ 
			\multicolumn{1}{l|}{} & \multicolumn{1}{c||}{HieraAgg$^\dagger$} 		
			& 9.4  & 34.77  	& 14.89  	& 33.13  & 33.51 \\ 
			\multicolumn{1}{l|}{} & \multicolumn{1}{c||}{TA$^\dagger$} 		
			& 8.7  & 35.14  	& 14.84  	& 33.26  & 33.32 \\ 
			\multicolumn{1}{l|}{} & \multicolumn{1}{c||}{deep MV-Transformer} 		
			& \textbf{10.8}  & \textbf{35.95}  & \textbf{15.71} & \textbf{33.90}   & \textbf{34.10} \\ 
			\multicolumn{1}{l|}{} & \multicolumn{1}{c||}{$\Delta$} 		
			& +1.7 & +0.57 & +1.44 & +0.75 & +0.49 \\ \hline 
			
			\multicolumn{1}{l|}{\multirow{6}{*}{Pri.}} &
			\multicolumn{1}{c||}{deep Transformer$^\dagger$} 		
			& 9.7 & 35.75 & 15.03 & 33.55 & 34.06 \\ 
			\multicolumn{1}{l|}{} &
			\multicolumn{1}{c||}{MLRF$^\dagger$} 		
			& 10.0  & 35.99  	& 15.0  	& 33.24  & \textbf{34.57}$^*$ \\ 
			\multicolumn{1}{l|}{} & \multicolumn{1}{c||}{HieraAgg$^\dagger$} 		
			& 8.9  & 35.03  	& 14.48  	& 32.57  & 33.82 \\ 
			\multicolumn{1}{l|}{} & \multicolumn{1}{c||}{TA$^\dagger$} 		
			& 9.0  & 34.63  	& 14.85  	& 33.78  & 33.88 \\ 
			\multicolumn{1}{l|}{} & \multicolumn{1}{c||}{deep MV-Transformer} 		
			& \textbf{10.8}$^*$  & \textbf{36.23}$^*$  & \textbf{15.74}$^*$ & \textbf{34.05} & \textbf{34.57}$^*$ \\ 
			\multicolumn{1}{l|}{} & \multicolumn{1}{c||}{$\Delta$} 		
			& +1.1 & +0.48 & +0.71 & +0.50 & +0.51 \\ 
			
			\bottomrule[1.pt]
		\end{tabular}
		
		\vspace{-0.5em}
		\caption{BLEU scores on five translation tasks. For (deep) Transformer, \texttt{Aux.}/\texttt{Pri.} denotes the independently trained model with $M_a$/$M$-layer encoder respectively. For (deep) MV-Transformer, \texttt{Aux.}/\texttt{Pri.} denotes the used view at inference time. $\Delta$ denotes the improved BLEU score over the Transformer baseline when using multi-view learning at the same encoder depth. $\dagger$ denotes our implementation. Boldface and $^*$ represent local and global best results, respectively.  All the MV-Transformer results are significantly better (p$<$0.01) than the Transformer counterparts, measured by paired bootstrap resampling \cite{koehn-2004-statistical}.}
		\label{table:main}
		\vspace{-1em}
	\end{center}
\end{table*}

\noindent 
In addition to Transformer, we also re-implemented three previously proposed models that incorporate multiple encoder layers: multi-layer representation fusion (MLRF)\cite{wang2018multi}, hierarchical aggregation (HieraAgg) \cite{dou2018exploiting}, and transparent attention (TA) \cite{bapna2018training}. 
Table \ref{table:main} shows the results of the five translation tasks on PostNorm and PreNorm. First, our MV-Transformer outperforms all baselines across the board. Specifically, for PostNorm models, with the helper of multi-view learning, both views can improve the Transformer baselines by about 0.4-1.5 BLEU points. Consistent improvements of 0.5-1.7 BLEU points are also obtained even in the stronger PreNorm baselines that benefit from the encoder's increased depth. And we achieve the new state-of-the-art of 10.8 and 36.23 on Ko$\rightarrow$En and De$\rightarrow$En, respectively \footnote{The previous state-of-the-art is 10.3 \cite{sennrich-zhang-2019-revisiting} on Ko$\rightarrow$En and 35.6 \cite{zhang-etal-2019-improving} on De$\rightarrow$En. }.
Note that these five tasks include both low-resource scenarios (Ko$\rightarrow$En) and rich-resource scenarios (En$\rightarrow$De), which indicates that our method has a good generalization for the scale of data size. 

\subsection{Compare to knowledge distillation and model ensemble}
\label{sec:cmp_kd_ensemble}

\begin{table}[tbp]
	\begin{center}
		\begin{tabular}{cc}
			\begin{minipage}[t]{.5\textwidth}
			\centering
			\begin{tabular}{c c c c }
				\toprule[1pt]
				
				\multicolumn{1}{c }{\textbf{Model} } &
				\multicolumn{1}{c }{\textbf{Speed} } &
				\multicolumn{1}{c }{\textbf{De$\rightarrow$En}} &
				\multicolumn{1}{c }{\textbf{Ro$\rightarrow$En} } \\
				\hline 
				
				Baseline (3L) & +0.3\% & 34.10 & 33.07 \\
				Baseline  & reference  & 34.77 & 33.20 \\
				Oneway-KD  & +0.8\% & 35.07 & 33.78 \\
				Seq-KD (3L) & +4.7\% & 34.13 & 32.45 \\
				Ensemble & -45.3\% & \textbf{35.77} & 34.15 \\
				MV (3L) & +1.6\% & 35.25 & 34.24 \\
				MV & +0.3\% & 35.49 & \textbf{34.45} \\
				
				\bottomrule[1pt]
			\end{tabular}
			\end{minipage}
			&
			\begin{minipage}[t]{.45\textwidth}
			\centering
			\begin{tabular}{c  c c  c }
				\toprule[1pt]
				
				\textbf{ID} & \multicolumn{1}{c }{\textbf{Gold}} &
				\multicolumn{1}{c }{\textbf{Dark}} &
				\multicolumn{1}{c }{\textbf{BLEU}} \\
				\hline 
				
				\#1 & $\surd$	& All 	& 35.49 \\
				
				\#2 & $\surd$	& - 	& 35.02 (-0.47) \\
				\#3 & -			& All	& 35.44 (-0.05) \\	 
				
				\#4 & -			& [1,100]	& 35.35 (-0.14) \\	 
				\#5 & -			& [100,]	& 34.79 (-0.70)\\	 
				
				\bottomrule[1pt]
			\end{tabular}
			\end{minipage} \\ 
			\begin{minipage}[t]{.5\textwidth}
				\vspace{-0.5em}
			\caption{Compare multi-view learning (MV) to oneway knowledge distillation (Oneway-KD), sequence-level knowledge distillation (Seq-KD) and model ensemble (Ensemble) on IWSLT'14 De$\rightarrow$En and WMT'16 En$\rightarrow$Ro test sets. Speed is the average of 5 runs with \texttt{batch}=32, \texttt{beam}=5.}
			\label{table:cmp_kd_ensemble}
			\vspace{-1.em}
			\end{minipage}
			&
			\begin{minipage}[t]{.45\textwidth}
			\vspace{-0.5em}
			\caption{Dark knowledge in our multi-view learning. \textit{Gold} is the ground-truth label, while \textit{Dark} denotes the set of all non-ground-truth labels. $[a, b]$ denotes the top-a to top-b predicted labels (exclude ground-truth).    }
			\label{table:dark}
			\vspace{-1.0em}
			\end{minipage} 
			
		\end{tabular}
	\end{center}
\end{table}

\noindent MV-Transformer can be thought of as consisting of two models: A large model as the primary view, and a small model (with shallower encoder) as the auxiliary view. Here we compare with the other three methods of integrating multiple models:

\begin{itemize}[]
	\item \textbf{Oneway-KD. }  Similar to Eq.~\ref{eq:mv_loss} but detach the teacher's prediction, i.e., gradients of the teacher's prediction is not tracked, posing a one-way transfer from primary view to auxiliary view.
	\item \textbf{Seq-KD. } Train the large model first and then translate the original training set by beam search to construct the distilled training set for the small model \cite{kim2016sequence}. 
	\item \textbf{Ensemble. } Independently train the two models and combine their predictions at inference time, e.g., by algorithmic average. 
\end{itemize}

Experiments are done on IWSLT'14 De$\rightarrow$En, where the small model has a 3-layer encoder. As shown in Table~\ref{table:cmp_kd_ensemble}, we can see that:
(1) \textit{Oneway-KD} suffers from severe degradation than \textit{MV} when detaching the primary view, which indicates that making mutual learning between the primary view and auxiliary view is critical;
(2) \textit{Seq-KD} is almost useless or even badly hurts the performance (vs. \textit{Baseline (3L)}), which is against the previous belief that \textit{Seq-KD} helps the small model a lot by learning from the teacher. We suspect that the reason is that our student's performance has already been closed to the teacher;
(3) \textit{Ensemble} can achieve significant performance improvement than a single model but at the cost of almost twice the slower decoding speed. However, our approach can achieve comparable or even better results than the model ensemble but maintain the decoding speed as a single model.

\section{Analysis}

\subsection{Why multi-view learning works?}
\label{sec:exp_why}

\begin{figure*}[t]
	\begin{center}
		\renewcommand\arraystretch{0.0}
		\begin{tabular}{cc}
			
			\subfloat[\footnotesize{6-layer encoder}]
			{
				\begin{tikzpicture}{baseline}
				\scriptsize{
					\begin{axis}[
					ylabel near ticks,
					width=.45\textwidth,
					height=.23\textwidth,
					legend style={at={(0.52,0.3)}, anchor=south east},
					xlabel={$\epsilon$},
					ylabel={\scriptsize{BLEU Score}},
					ylabel style={yshift=-0em},xlabel style={yshift=0.0em},
					yticklabel style={/pgf/number format/precision=0,/pgf/number format/fixed zerofill},
					xmin=0,xmax=1.0,xtick={.0,.1,.2,.3,.4,.5,.6,.7,.8,.9,1.0},
					xmajorgrids,
					ymajorgrids,
					axis lines*=left,
					legend style={yshift=-12pt, legend plot pos=left,cells={anchor=west}}
					]
					
					\addplot[red,mark=otimes*,line width=0.5pt] coordinates { (0., 34.77) (.1,34.58) (.2,34.18) (.3,33.31) (.4,31.47) (.5,27.74) (.6,21.47) (.7,13.44) (.8, 6.36) (.9, 2.71) (1.0,1.22)};
					\addlegendentry{transformer}
					
					\addplot[blue,mark=square,line width=0.5pt] coordinates { (0., 35.49) (.1,35.49) (.2,35.18) (.3,34.62) (.4,33.80) (.5,32.93) (.6,30.74) (.7,27.40) (.8, 22.08) (.9, 16.50) (1.0,10.29)};
					\addlegendentry{mv-transformer}

					\end{axis}
				}
				\label{fig:noise_6l}
				\end{tikzpicture}
			}
			&
			\subfloat [\footnotesize{12-layer encoder}]
			{
				\begin{tikzpicture}{baseline}
				\scriptsize{
					\begin{axis}[
					ylabel near ticks,
					width=.45\textwidth,
					height=.23\textwidth,
					legend style={at={(0.52,0.3)}, anchor=south east},
					xlabel={$\epsilon$},
					ylabel={\scriptsize{BLEU Score}},
					ylabel style={yshift=-0em},xlabel style={yshift=0.0em},
					yticklabel style={/pgf/number format/precision=0,/pgf/number format/fixed zerofill},
					ymin=28,ymax=37.5,ytick={36, 34, 32, 30, 28},
					xmin=0,xmax=1.0,xtick={.0,.1,.2,.3,.4,.5,.6,.7,.8,.9,1.0},
					xmajorgrids,
					ymajorgrids,
					axis lines*=left,
					legend style={yshift=-12pt, legend plot pos=left,cells={anchor=west}}
					]
					
					\addplot[red,mark=otimes*,line width=0.5pt] coordinates { (0., 35.75) (.1,35.81) (.2,35.56) (.3,35.50) (.4,35.20) (.5,34.89) (.6,34.28) (.7,33.79) (.8,32.74) (.9, 31.28) (1.0,28.89)};
					\addlegendentry{transformer}
					
					\addplot[blue,mark=square,line width=0.5pt] coordinates { (0., 36.23) (.1,36.30) (.2,36.07) (.3,35.93) (.4,35.85) (.5,35.47) (.6,35.16) (.7,34.53) (.8, 33.99) (.9,32.83) (1.0,30.98)};
					\addlegendentry{mv-transformer}
					
					\end{axis}
				}
				\label{fig:noise_12l}
				\end{tikzpicture}
			}
			
		\end{tabular}
	\end{center}
	
	\begin{center}
		\vspace{-0.5em}
		\caption{ BLEU scores on IWSLT'14 De$\rightarrow$En test set w.r.t injecting noises sampled from $\mathcal{N}(0, \epsilon)$ on the encoding representation. (a) 6-layer encoder with PostNorm; (b) 12-layer encoder with PreNorm.}
		\label{fig:noise}
		\vspace{-1.0em}
	\end{center}
\end{figure*}
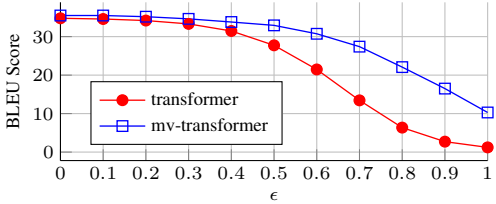
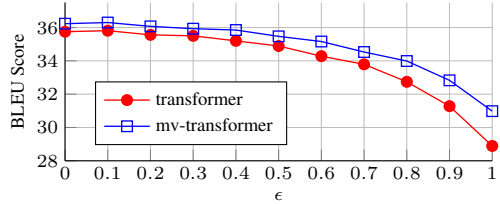

\begin{enumerate}[leftmargin=0pt]
	
	\item[] \textbf{Robustness to encoding noises. }  In \cref{sec:mv_discuss}, we suppose that MV-Transformer is less sensitive to noise encoding representations due to the introduction of the auxiliary view. To verify it, we add random Gaussian noises sampled from $\mathcal{N}(0, \epsilon)$ into the normalized input in the last layer normalization of the encoder\footnote{Original layer normalization is $y=g \odot N(x)+b$, while our noised version is $y=g \odot \big(N(x)+\epsilon\big)+b$. Our purpose is to avoid different scales of $g$ and $b$ between models.}. As shown in Figure \ref{fig:noise}, we can see that while both models degrade performance along with stronger noises, MV-Transformer is less sensitive, e.g., the maximum gap is 15.72/2.09 when $\epsilon$=0.8/1.0 for the 6/12-layer encoder respectively. It indicates that our model has a better generalization even if the test distribution is largely different from the training distribution.  
	We also observed that the PreNorm-style Transformer is less sensitive than the PostNorm counterpart, e.g., when $\epsilon$=1.0, the PostNorm Transformer decreases 33.55 BLEU points, while the PreNorm one only decreases 6.86. 
	
	\item[] \textbf{Dark knowledge. } As shown in Table~\ref{table:dark}, we study the effect of dark knowledge in our multi-view learning. First, we test the case where we only use gold knowledge (ground-truth label, \#2) and dark knowledge (non-ground truth labels, \#3). Obviously, without the help of dark knowledge, multi-view learning fails to boost the performance. Going further along this line, we study which part of dark knowledge is the most important (\#4-5). Specifically, we split all non-gold labels into two parts from high to low according to their probability: [1,100] and [100,]. We can see that our approach's success lies in those non-gold labels at the top, not the long tail part.
	
\end{enumerate}

\subsection{Hyperparameter sensitivity}
\label{sec:exp_hp}

\begin{figure}[tbp]
	\captionsetup[subfigure]{position=top}
	\begin{center}
		\begin{tabular}{cc}
			
			\begin{minipage}[t]{.45\textwidth}
				
				\begin{tabular}[t]{cc}
					\subfloat[\footnotesize{6-layer encoder}]
					{
						\begin{tikzpicture}{baseline}
						\scriptsize{
							\begin{axis}[
							ylabel near ticks,
							width=.5\textwidth,
							height=.47\textwidth,
							xlabel={$M_a$},
							ylabel={\scriptsize{BLEU Score}},
							ylabel style={yshift=-0em},xlabel style={yshift=0.0em},
							yticklabel style={/pgf/number format/precision=1,/pgf/number format/fixed zerofill},
							xmin=0,xmax=5,xtick={0,1,2,3,4,5},
							]
							
							\addplot[red,mark=otimes*,line width=0.5pt] coordinates { (0, 34.77) (1,35.15) (2,35.27) (3,35.49) (4,35.33) (5,35.42)};
							
							\end{axis}
						}
						\label{fig:view_pos_de2en}
						\end{tikzpicture}
					}
					&
					\subfloat [\footnotesize{12-layer encoder}]
					{
						\begin{tikzpicture}{baseline}
						\scriptsize{
							\begin{axis}[
							ylabel near ticks,
							width=.5\textwidth,
							height=.47\textwidth,
							xlabel={$M_a$},
							ylabel={\scriptsize{BLEU Score}},
							ylabel style={yshift=-0em},xlabel style={yshift=0.0em},
							yticklabel style={/pgf/number format/precision=1,/pgf/number format/fixed zerofill},
							xmin=0,xmax=10,xtick={0,2,4,6,8,10},
							]
							
							\addplot[red,mark=otimes*,line width=0.5pt] coordinates {(0,35.75) (2,35.87) (4,36.1) (6,36.23) (8,36.2) (10,36.37)};
							
							\end{axis}
						}
						\label{fig:view_pos_ro2en}
						\end{tikzpicture}
					}
				\end{tabular}
			\end{minipage}
			&
			\begin{minipage}[t]{.45\textwidth}
				\captionsetup[subfigure]{labelformat=empty}
				\subfloat []
				{
					\begin{tikzpicture}{baseline}
					\scriptsize{
						\begin{axis}[
						ylabel near ticks,
						width=.95\textwidth,
						height=.48\textwidth,
						legend style={at={(0.8, 0.5)}, anchor=north east},
						xlabel={$\alpha$},
						ylabel={\scriptsize{BLEU Score}},
						ylabel style={yshift=-0em},xlabel style={yshift=0.0em},
						yticklabel style={/pgf/number format/precision=1,/pgf/number format/fixed zerofill},
						ymin=34.5,ymax=36.0,
						xmin=0.0,xmax=0.8, xtick={0, 0.1, 0.2, 0.3, 0.4, 0.5, 0.6, 0.7, 0.8},
						xmajorgrids,
						ymajorgrids,
						legend style={legend plot pos=left,cells={anchor=west}}
						]
						
						\addplot[red,mark=otimes*,line width=0.5pt] coordinates {(0.0, 35.11) (0.1,35.11) (0.2, 35.26) (0.3,35.49) (0.4, 35.69) (0.5, 35.6) (0.6,35.46) (0.7,35.25) (0.8,34.64) (0.9,29.8)};
						\addlegendentry{\scriptsize{primary}}
						
						\addplot[blue,mark=square,line width=0.5pt] coordinates {(0.0,34.56) (0.1,34.6) (0.2,34.96) (0.3,35.25) (0.4, 35.43) (0.5, 35.48) (0.6,35.46) (0.7,35.25) (0.8,34.72) (0.9,28.81)};
						\addlegendentry{\scriptsize{auxiliary}}
						
						\end{axis}
					}
					\label{fig:kl_alpha_de2en}
					\end{tikzpicture}
				}
				
			\end{minipage} \\ 
		
			\begin{minipage}[t]{.45\textwidth}
				\vspace{-0.5em}
				\caption{ BLEU scores against the position of auxiliary view ($M_a$) on IWSLT'14 De$\rightarrow$En. $M_a$=0: no multi-view. }
				\label{fig:view_pos}
				\vspace{-1.0em}
			\end{minipage} &
		
			\begin{minipage}[t]{.45\textwidth}
				\vspace{-0.5em}
				\caption{BLEU scores against $\alpha$ for multi-view learning on  IWSLT'14 De$\rightarrow$En test set. $M_a$=3, $M$=6. }
				\label{fig:kl_alpha}
				\vspace{-1.0em}
			\end{minipage}
			
		\end{tabular}
	\end{center}
\end{figure}

\begin{enumerate}[leftmargin=0pt]
	
	\item[] \textbf{Position of auxiliary view. }  In Figure \ref{fig:view_pos}, we plot the BLEU score curves against the auxiliary view position $L_a$ on IWSLT'14 De$\rightarrow$En. In general, we can obtain better performance when $L_a$ is closer to $L_e$, but this is not always true. It is intuitive: When $L_a \ll L_e$, the auxiliary view is numerically far different from the primary view, which is difficult for a partially shared decoder to learn. On the contrary, if $L_a$ is too close to the top, the slight difference may not bring too many learnable signals. In this work, we use the middle layer as the auxiliary view, while it should be noted that we could obtain better results if we tune $L_a$ more carefully (e.g., $L_a$=10 vs. $L_a$=6 in the 12-layer encoder).
	
	\item[] \textbf{Interpolation coefficient. }  In Figure \ref{fig:kl_alpha}, we show the curve of BLEU score against the hyperparameter $\alpha$ on IWSLT'14 De$\rightarrow$En test set. First of all, we can see that the BLEU score is improved along with the increase of $\alpha$, while it starts to decrease when $\alpha$ is too large (e.g., $\alpha >$ 0.6). In particular, we failed to train the model when $\alpha$=1.0. On the other hand, a large $\alpha$ can reduce the gap between the two views as expected. We note that even though it is difficult to know the optimal $\alpha$ in advance, we empirically found that $\alpha \in [0.3, 0.5]$ is robust across these distinct tasks.
	
\end{enumerate}

\subsection{Transparency to network architecture}

\begin{table}[tbp]
	\begin{center}
		\begin{tabular}{cc}
				\begin{minipage}[t]{.45\textwidth}
				
				\begin{tabular}[t]{c  c c }
					\toprule[1pt]
					
					\multicolumn{1}{c }{\textbf{Model} } &
					\multicolumn{1}{c }{\textbf{De$\rightarrow$En}} &
					\multicolumn{1}{c }{} \\
					\hline 
					
					DynamicConv \cite{wu2019pay} & 35.2 & ~ \\
					DynamicConv$^\dagger$ (4 layers) & 35.13 & ~ \\
					DynamicConv$^\dagger$ (7 layers) & 35.21 & ~ \\
					MV-DynamicConv (auxiliary) & 35.59 & ~ \\
					MV-DynamicConv (primary) & \textbf{35.79} & ~ \\
					
					\bottomrule[1pt]
				\end{tabular}
				\end{minipage}
				&
				\begin{minipage}[t]{.45\textwidth}
				\begin{tabular}[t]{c c  c  c }
					\toprule[1pt]
					
					\multicolumn{1}{c }{\textbf{ID}} & \multicolumn{1}{c }{\textbf{Model}} & \multicolumn{1}{c }{\textbf{DE$\rightarrow$EN}} &
					\multicolumn{1}{c }{\textbf{RO$\rightarrow$EN}} \\
					\hline 
					
					\#1 & Baseline & 34.77 & 33.20 \\
					\#2 & MV-3-6 & 35.49 & 34.45 \\
					\#3 & MV-3-6 (shared)  & 35.51 & 33.93 \\
					\#4 & MV-6-6 & 35.05 & 33.86 \\ 
					\bottomrule[1pt]
				\end{tabular} 
			\end{minipage} \\ 
				\begin{minipage}[t]{.45\textwidth}
				\vspace{-0.5em}
				\caption{Apply multi-view learning to DynamicConv with $M_a$=4 and $M$=7 on IWSLT'14 De$\rightarrow$En test set. $\dagger$ denotes our implementation.}
				\label{table:other_model}
				\vspace{-1.em}
			\end{minipage}
			 &
			\begin{minipage}[t]{.45\textwidth}
				\vspace{-0.5em}
				\caption{Ablation study. \texttt{MV-\#1-\#2} denotes $M_a$=\#1 and $M$=\#2, \texttt{(shared)} indicates the use of shared CAN sublayers in decoder. 
				}
				\label{table:ablation2}
				\vspace{-1.0em}
			\end{minipage} 

		\end{tabular}
	\end{center}
\end{table}

In addition to the Transformer, we also test our method on the recently proposed DynamicConv \cite{wu2019pay}. Original DynamicConv is composed of a 7-layer encoder and a 6-layer decoder. Our method takes the topmost layer as the primary view as before and uses the 4-th layer as the auxiliary view. The results on IWSLT'14 De$\rightarrow$En task are listed in Table \ref{table:other_model}. It can be seen that DynamicConv with multi-view is stably higher than that of a single-view model by about 0.5 BLEU score, which indicates that our method is transparent to network architecture and has the potential to be widely used.

\subsection{Ablation study}
\label{sec:exp_ablation}

\noindent We did ablation studies to understand the effects of (a) separate CANs and (b)  using a lower encoder layer as auxiliary view. Experimental results are listed in Table~\ref{table:ablation2}. We can see that:
(1) Under almost the same parameter size, sharing CANs (\#3) obtains +0.7 BLEU in both tasks compared to the baseline (\#1), which indicates the improvement comes from our multi-view training instead of the increased parameters;
(2) Using separate CANs is more helpful than sharing CANs when the size of training data is large enough (\#3 vs. \#2);
(3) Thanks to separate CANs, the decoder can obtain distinguishable context representations even if the auxiliary view is the same as the primary view  (\#4 vs. \#1);
(4) The auxiliary view with a high layer (\#4, $M_a$=6) performs worse than that of a low layer (\#2, $M_a$=3), which strongly indicates the diversity between views is more important than the quality of the auxiliary view.

\section{Related Work}


\begin{enumerate}[leftmargin=0pt]
	
	\item[] \textbf{Multi-view learning. }
	\noindent 
	In multi-view learning, one of the most fundamental problems is view construction. Most previous works study random sampling in the feature spaces \cite{ho1998the}, feature vector transformation by reshaping \cite{wang2011a}. For natural language processing, \citet{Matsubara:2005:MSL:1565899.1565913} obtain the multiple views of one document by taking different grams as terms in the bag-of-word model. Perhaps the most related work in this topic is \citet{clark-etal-2018-semi}, which randomly mask input tokens to generate different sequences. 
	Different from \citet{clark-etal-2018-semi}, we take the off-the-shelf outputs of the encoder layers as views which is more general for multi-layer networks without any construction cost. 
	
	\item[] \textbf{Consistency regularization. }
	Knowledge distillation (KD) is a typical application of consistency regularization, which achieves knowledge transfer by letting the student model imitate the teacher model \cite{hinton2015distilling}. 
	There are many ways to construct the student model. For example,
	the student is the peer model as the teacher in \citet{ying2018DML}, and \citet{lan2018knowledge} take one branch as the student in their multi-branch network architecture. As for us, our student model consists of a shallow teacher network in a partially shared manner.
	Another important application scenario of consistency regularization is semi-supervised learning, such as Temporal Ensembling \cite{tempensemble}, Mean Teacher \cite{meanteacher}, Virtual Adversarial Training \cite{vat2017} etc. However, our method works in supervised learning without the requirement of unlabeled data.
	
\end{enumerate}

\section{Conclusion}
\noindent We studied to incorporate different encoder layers through multi-view learning in neural machine translation. In addition to the primary view from the topmost layer, the proposed model introduces an auxiliary view from an intermediate encoder layer and encourages the transfer of knowledge between the two views. Our method is agnostic to network architecture and can maintain the same inference speed as the original model. We tested our method on five translation tasks with multiple strong baselines: Transformer, deep Transformer, and DynamicConv. Experimental results show that our multi-view learning method can stably outperform the baseline models. Our models have achieved new state-of-the-art results in Ko$\rightarrow$En and IWSLT'14 De$\rightarrow$En tasks. 

\section*{Acknowledgements}
This work was supported in part by the National Science Foundation of China (Nos. 61876035 and 61732005), the National Key R\&D Program of China (No. 2019QY1801). Yue Zhang was funded by the machine translation project of Alibaba DAMO Academy.

\bibliographystyle{coling}
\bibliography{coling2020}

\begin{thebibliography}{}

\bibitem[\protect\citename{Ba \bgroup et al.\egroup }2016]{ba16layer}
Lei~Jimmy Ba, Ryan Kiros, and Geoffrey~E. Hinton.
\newblock 2016.
\newblock Layer normalization.
\newblock {\em CoRR}, abs/1607.06450.

\bibitem[\protect\citename{Bahdanau \bgroup et al.\egroup
  }2015]{bahdanau2015neural}
Dzmitry Bahdanau, Kyunghyun Cho, and Yoshua Bengio.
\newblock 2015.
\newblock Neural machine translation by jointly learning to align and
  translate.
\newblock In {\em 3rd International Conference on Learning Representations,
  ICLR 2015}.

\bibitem[\protect\citename{Bapna \bgroup et al.\egroup
  }2018]{bapna2018training}
Ankur Bapna, Mia Chen, Orhan Firat, Yuan Cao, and Yonghui Wu.
\newblock 2018.
\newblock Training deeper neural machine translation models with transparent
  attention.
\newblock In {\em Proceedings of the 2018 Conference on Empirical Methods in
  Natural Language Processing}, pages 3028--3033.

\bibitem[\protect\citename{Cheng \bgroup et al.\egroup
  }2019]{cheng-etal-2019-robust}
Yong Cheng, Lu~Jiang, and Wolfgang Macherey.
\newblock 2019.
\newblock Robust neural machine translation with doubly adversarial inputs.
\newblock In {\em Proceedings of the 57th Annual Meeting of the Association for
  Computational Linguistics}, pages 4324--4333, Florence, Italy, July.
  Association for Computational Linguistics.

\bibitem[\protect\citename{Clark \bgroup et al.\egroup
  }2018]{clark-etal-2018-semi}
Kevin Clark, Minh-Thang Luong, Christopher~D. Manning, and Quoc Le.
\newblock 2018.
\newblock Semi-supervised sequence modeling with cross-view training.
\newblock In {\em Proceedings of the 2018 Conference on Empirical Methods in
  Natural Language Processing}, pages 1914--1925, Brussels, Belgium,
  October-November.

\bibitem[\protect\citename{Dou \bgroup et al.\egroup }2018]{dou2018exploiting}
Zi-Yi Dou, Zhaopeng Tu, Xing Wang, Shuming Shi, and Tong Zhang.
\newblock 2018.
\newblock Exploiting deep representations for neural machine translation.
\newblock In {\em Proceedings of the 2018 Conference on Empirical Methods in
  Natural Language Processing}, pages 4253--4262.

\bibitem[\protect\citename{He \bgroup et al.\egroup }2016]{he2016deep}
Kaiming He, Xiangyu Zhang, Shaoqing Ren, and Jian Sun.
\newblock 2016.
\newblock Deep residual learning for image recognition.
\newblock In {\em Proceedings of the IEEE conference on computer vision and
  pattern recognition}, pages 770--778.

\bibitem[\protect\citename{He \bgroup et al.\egroup }2018]{he2018layer}
Tianyu He, Xu~Tan, Yingce Xia, Di~He, Tao Qin, Zhibo Chen, and Tie-Yan Liu.
\newblock 2018.
\newblock Layer-wise coordination between encoder and decoder for neural
  machine translation.
\newblock In {\em Advances in Neural Information Processing Systems}, pages
  7955--7965.

\bibitem[\protect\citename{Hinton \bgroup et al.\egroup
  }2015]{hinton2015distilling}
Geoffrey Hinton, Oriol Vinyals, and Jeff Dean.
\newblock 2015.
\newblock Distilling the knowledge in a neural network.
\newblock {\em arXiv preprint arXiv:1503.02531}.

\bibitem[\protect\citename{Ho}1998]{ho1998the}
Tin~Kam Ho.
\newblock 1998.
\newblock The random subspace method for constructing decision forests.
\newblock {\em IEEE Transactions on Pattern Analysis and Machine Intelligence},
  20(8):832--844.

\bibitem[\protect\citename{Kim and Rush}2016]{kim2016sequence}
Yoon Kim and Alexander~M Rush.
\newblock 2016.
\newblock Sequence-level knowledge distillation.
\newblock In {\em Proceedings of the 2016 Conference on Empirical Methods in
  Natural Language Processing}, pages 1317--1327.

\bibitem[\protect\citename{Koehn}2004]{koehn-2004-statistical}
Philipp Koehn.
\newblock 2004.
\newblock Statistical significance tests for machine translation evaluation.
\newblock In {\em Proceedings of the 2004 Conference on Empirical Methods in
  Natural Language Processing}, pages 388--395, Barcelona, Spain, July.
  Association for Computational Linguistics.

\bibitem[\protect\citename{Laine and Aila}2017]{tempensemble}
Samuli Laine and Timo Aila.
\newblock 2017.
\newblock Temporal ensembling for semi-supervised learning.
\newblock In {\em Proc. International Conference on Learning Representations
  (ICLR)}.

\bibitem[\protect\citename{Lan \bgroup et al.\egroup }2018]{lan2018knowledge}
Xu~Lan, Xiatian Zhu, and Shaogang Gong.
\newblock 2018.
\newblock Knowledge distillation by on-the-fly native ensemble.
\newblock In {\em Proceedings of the 32nd International Conference on Neural
  Information Processing Systems}, pages 7528--7538. Curran Associates Inc.

\bibitem[\protect\citename{Matsubara \bgroup et al.\egroup
  }2005]{Matsubara:2005:MSL:1565899.1565913}
Edson~Takashi Matsubara, Maria~Carolina Monard, and Gustavo E. A. P.~A.
  Batista.
\newblock 2005.
\newblock Multi-view semi-supervised learning: An approach to obtain different
  views from text datasets.
\newblock In {\em Proceedings of the 2005 Conference on Advances in Logic Based
  Intelligent Systems}, pages 97--104.

\bibitem[\protect\citename{Miyato \bgroup et al.\egroup }2017]{vat2017}
Takeru Miyato, Andrew~M. Dai, and Ian~J. Goodfellow.
\newblock 2017.
\newblock Adversarial training methods for semi-supervised text classification.
\newblock In {\em 5th International Conference on Learning Representations,
  {ICLR} 2017}.

\bibitem[\protect\citename{Peters \bgroup et al.\egroup }2018]{peters2018deep}
Matthew Peters, Mark Neumann, Mohit Iyyer, Matt Gardner, Christopher Clark,
  Kenton Lee, and Luke Zettlemoyer.
\newblock 2018.
\newblock Deep contextualized word representations.
\newblock In {\em Proceedings of the 2018 Conference of the North American
  Chapter of the Association for Computational Linguistics: Human Language
  Technologies, Volume 1 (Long Papers)}, volume~1, pages 2227--2237.

\bibitem[\protect\citename{Raganato and
  Tiedemann}2018]{raganato-tiedemann-2018-analysis}
Alessandro Raganato and J{\"o}rg Tiedemann.
\newblock 2018.
\newblock An analysis of encoder representations in transformer-based machine
  translation.
\newblock In {\em Proceedings of the 2018 {EMNLP} Workshop {B}lackbox{NLP}:
  Analyzing and Interpreting Neural Networks for {NLP}}, pages 287--297,
  Brussels, Belgium, November.

\bibitem[\protect\citename{Sennrich and
  Zhang}2019]{sennrich-zhang-2019-revisiting}
Rico Sennrich and Biao Zhang.
\newblock 2019.
\newblock {Revisiting Low-Resource Neural Machine Translation: A Case Study}.
\newblock In {\em {Proceedings of the 57th Conference of the Association for
  Computational Linguistics}}, pages 211--221, Florence, Italy, July.

\bibitem[\protect\citename{{Srivastava} \bgroup et al.\egroup
  }2014]{srivastava2014dropout}
Nitish {Srivastava}, Geoffrey {Hinton}, Alex {Krizhevsky}, Ilya {Sutskever},
  and Ruslan {Salakhutdinov}.
\newblock 2014.
\newblock Dropout: a simple way to prevent neural networks from overfitting.
\newblock {\em Journal of Machine Learning Research}, 15(1):1929--1958.

\bibitem[\protect\citename{Szegedy \bgroup et al.\egroup
  }2016]{szegedy2016rethinking}
Christian Szegedy, Vincent Vanhoucke, Sergey Ioffe, Jon Shlens, and Zbigniew
  Wojna.
\newblock 2016.
\newblock Rethinking the inception architecture for computer vision.
\newblock In {\em Proceedings of the IEEE conference on computer vision and
  pattern recognition}, pages 2818--2826.

\bibitem[\protect\citename{Tarvainen and Valpola}2017]{meanteacher}
Antti Tarvainen and Harri Valpola.
\newblock 2017.
\newblock Mean teachers are better role models: Weight-averaged consistency
  targets improve semi-supervised deep learning results.
\newblock In I.~Guyon, U.~V. Luxburg, S.~Bengio, H.~Wallach, R.~Fergus,
  S.~Vishwanathan, and R.~Garnett, editors, {\em Advances in Neural Information
  Processing Systems 30}, pages 1195--1204. Curran Associates, Inc.

\bibitem[\protect\citename{Vaswani \bgroup et al.\egroup
  }2017]{vaswani2017attention}
Ashish Vaswani, Noam Shazeer, Niki Parmar, Jakob Uszkoreit, Llion Jones,
  Aidan~N Gomez, {\L}ukasz Kaiser, and Illia Polosukhin.
\newblock 2017.
\newblock Attention is all you need.
\newblock In {\em Advances in Neural Information Processing Systems}, pages
  6000--6010.

\bibitem[\protect\citename{Wang \bgroup et al.\egroup }2011]{wang2011a}
Zhe Wang, Songcan Chen, and Daqi Gao.
\newblock 2011.
\newblock A novel multi-view learning developed from single-view patterns.
\newblock {\em Pattern Recognition}, 44(10):2395--2413.

\bibitem[\protect\citename{Wang \bgroup et al.\egroup }2018]{wang2018multi}
Qiang Wang, Fuxue Li, Tong Xiao, Yanyang Li, Yinqiao Li, and Jingbo Zhu.
\newblock 2018.
\newblock Multi-layer representation fusion for neural machine translation.
\newblock In {\em Proceedings of the 27th International Conference on
  Computational Linguistics}, pages 3015--3026.

\bibitem[\protect\citename{Wang \bgroup et al.\egroup }2019a]{wang2019learning}
Qiang Wang, Bei Li, Tong Xiao, Jingbo Zhu, Changliang Li, Derek~F. Wong, and
  Lidia~S. Chao.
\newblock 2019a.
\newblock Learning deep transformer models for machine translation.
\newblock In {\em Proceedings of the 57th Annual Meeting of the Association for
  Computational Linguistics}, pages 1810--1822, Florence, Italy, July.

\bibitem[\protect\citename{Wang \bgroup et al.\egroup
  }2019b]{wang-etal-2019-exploiting}
Xing Wang, Zhaopeng Tu, Longyue Wang, and Shuming Shi.
\newblock 2019b.
\newblock Exploiting sentential context for neural machine translation.
\newblock In {\em Proceedings of the 57th Annual Meeting of the Association for
  Computational Linguistics}, pages 6197--6203, Florence, Italy, July.

\bibitem[\protect\citename{Wu \bgroup et al.\egroup }2016]{wu2016google}
Yonghui Wu, Mike Schuster, Zhifeng Chen, Quoc~V Le, Mohammad Norouzi, Wolfgang
  Macherey, Maxim Krikun, Yuan Cao, Qin Gao, Klaus Macherey, et~al.
\newblock 2016.
\newblock Google's neural machine translation system: Bridging the gap between
  human and machine translation.
\newblock {\em arXiv preprint arXiv:1609.08144}.

\bibitem[\protect\citename{Wu \bgroup et al.\egroup }2019]{wu2019pay}
Felix Wu, Angela Fan, Alexei Baevski, Yann~N. Dauphin, and Michael Auli.
\newblock 2019.
\newblock Pay less attention with lightweight and dynamic convolutions.
\newblock In {\em 7th International Conference on Learning Representations,
  {ICLR} 2019}.

\bibitem[\protect\citename{Xu \bgroup et al.\egroup }2013]{xu2013survey}
Chang Xu, Dacheng Tao, and Chao Xu.
\newblock 2013.
\newblock A survey on multi-view learning.
\newblock {\em arXiv preprint arXiv:1304.5634}.

\bibitem[\protect\citename{Zhang \bgroup et al.\egroup }2018]{ying2018DML}
Ying Zhang, Tao Xiang, Timothy~M. Hospedales, and Huchuan Lu.
\newblock 2018.
\newblock Deep mutual learning.
\newblock In {\em CVPR}.

\bibitem[\protect\citename{Zhang \bgroup et al.\egroup
  }2019]{zhang-etal-2019-improving}
Biao Zhang, Ivan Titov, and Rico Sennrich.
\newblock 2019.
\newblock Improving deep transformer with depth-scaled initialization and
  merged attention.
\newblock In {\em Proceedings of the 2019 Conference on Empirical Methods in
  Natural Language Processing and the 9th International Joint Conference on
  Natural Language Processing (EMNLP-IJCNLP)}, pages 898--909, Hong Kong,
  China, November. Association for Computational Linguistics.

\end{thebibliography}

\appendix

\section*{Appendix A.   Hyper-parameters setting}

\noindent Table~\ref{table:hparam} shows the hyper-parameter list used in our experiments. It can be seen that these training hyper-parameters change across tasks according to model architecture, data scale, and learning difficulty. From the perspective of the model, the Transformer and deep Transformer uses layer normalization in post-norm/pre-norm form, respectively. When using a deep Transformer, we use a deeper encoder and double the batch size to reduce gradient variance with a larger learning rate for fast divergence. Also, we half the number of updates to guarantee the same quantity of training dataset seen for a fair comparison. Note that the decoder depth keeps the same.

\begin{table*}[h]
	\begin{center}
		\renewcommand\arraystretch{1}
		\begin{tabular}{l l| c  c  c  c  c}
			\toprule[1pt]
			
			\multicolumn{1}{c |}{\textbf{Model}} &
			\multicolumn{1}{c ||}{\textbf{Hyperparameter}} &
			
			\multicolumn{1}{c }{\textbf{Ko-En}} & 
			\multicolumn{1}{c }{\textbf{De-En}} &
			\multicolumn{1}{c }{\textbf{Tr-En}} &
			\multicolumn{1}{c }{\textbf{Ro-En}} &
			\multicolumn{1}{c }{\textbf{En-De}} \\

			\hline \hline
			\multicolumn{1}{l|}{\multirow{9}{*}{(MV-)Transformer}} &
			\multicolumn{1}{c||}{encoder layer} 		
			& 3  & 6  	& 6  	& 6   & 6  \\ 
			
			\multicolumn{1}{l|}{} & \multicolumn{1}{c||}{decoder layer} 		
			& 3  & 6  	& 6  	& 6  & 6 \\ 
			
			\multicolumn{1}{l|}{} & \multicolumn{1}{c||}{batch size} 		
			&  4096	& 4096 	& 4096 	& 4096*8 & 4096*8 \\ 
			
			\multicolumn{1}{l|}{} & \multicolumn{1}{c||}{learning rate} 		
			&  0.0003	& 0.0005 	& 0.0002 	& 0.0005 & 0.0007 \\ 
			
			\multicolumn{1}{l|}{} &
			\multicolumn{1}{c||}{warmup} 		
			& 16k  	& 4k 	& 16k  	& 4k  & 4k  \\ 
			
			\multicolumn{1}{l|}{} &
			\multicolumn{1}{c||}{update} 		
			&   100k	& 50k  	& 100k  	& 100k  & 100k  \\ 
			
			\multicolumn{1}{l|}{} &
			\multicolumn{1}{c||}{dropout} 		
			&  0.3 	& 0.3  	& 0.3 	& 0.3  & 0.1  \\ 
			
			\multicolumn{1}{l|}{} &
			\multicolumn{1}{c||}{attention dropout} 		
			& 0.1	& 0.0  	& 0.1 	& 0.0  & 0.0  \\ 
			
			\multicolumn{1}{l|}{} &
			\multicolumn{1}{c||}{$\alpha$} 		
			& 0.5  	& 0.3  	& 0.3  	& 0.4  & 0.4  \\ 
			
			\hline \hline
			\multicolumn{1}{l|}{\multirow{9}{*}{deep (MV-)Transformer}} &
			\multicolumn{1}{c||}{encoder layer} 		
			& 6  & 12  	& 12  	& 12   & 12  \\ 
			
			\multicolumn{1}{l|}{} & \multicolumn{1}{c||}{decoder layer} 		
			& 3  & 6  	& 6  	& 6  & 6 \\ 
			
			\multicolumn{1}{l|}{} & \multicolumn{1}{c||}{batch size} 		
			&  4096	& 4096*2 	& 4096*2 	& 4096*16 & 4096*16 \\ 
			
			\multicolumn{1}{l|}{} & \multicolumn{1}{c||}{learning rate} 		
			&  0.0004	& 0.0015 	& 0.001 	& 0.0015 & 0.002 \\ 
			
			\multicolumn{1}{l|}{} &
			\multicolumn{1}{c||}{warmup} 		
			& 16k  	& 16k 	& 16k  	& 16k  & 16k  \\ 
			
			\multicolumn{1}{l|}{} &
			\multicolumn{1}{c||}{update} 		
			&   100k	& 25k  	& 50k  	& 50k  & 50k  \\ 
			
			\multicolumn{1}{l|}{} &
			\multicolumn{1}{c||}{dropout} 		
			&  0.3 	& 0.3  	& 0.3 	& 0.3  & 0.1  \\ 
			
			\multicolumn{1}{l|}{} &
			\multicolumn{1}{c||}{attention dropout} 		
			& 0.1	& 0.1  	& 0.1 	& 0.1  & 0.1  \\ 
			
			\multicolumn{1}{l|}{} &
			\multicolumn{1}{c||}{$\alpha$} 		
			& 0.5  	& 0.3  	& 0.3  	& 0.4  & 0.3 \\ 
			
			\bottomrule[1.pt]
		\end{tabular}
		
		\vspace{-0.0em}
		\caption{Hyper-parameter list for (MV-)Transformer and deep (MV-)Transformer.}
		\label{table:hparam}
		\vspace{-0.5em}
	\end{center}
\end{table*}	

\end{document}